\pdfoutput=1

\documentclass[11pt]{article}

\usepackage[preprint]{acl}

\usepackage{times}
\usepackage{latexsym}

\usepackage[T1]{fontenc}

\usepackage[utf8]{inputenc}

\usepackage{microtype}

\usepackage{inconsolata}

\usepackage{multirow}
\usepackage{adjustbox}
\usepackage{linguex}
\usepackage{makecell}
\usepackage{amsmath}
\usepackage[breakable]{tcolorbox}
\usepackage{CJKutf8}

\title{Evaluating the Performance of Large Language Models on \\GAOKAO Benchmark}

\begin{document}

\author{
Xiaotian Zhang\textsuperscript{1,}\footnotemark[1] , Chunyang Li\textsuperscript{1,}\footnotemark[1] , Yi Zong\textsuperscript{1,}\footnotemark[1] , Zhengyu Ying\textsuperscript{2}, Liang He\footnotemark[2] , Xipeng Qiu\footnotemark[2]\\
\textbf{Tianxiang Sun}, \textbf{Peng Li}, 
\textbf{Shiqiao Meng}, \textbf{Yanjun Zheng}, \textbf{Jun Zhan}, \\
\textbf{Zhangyue Yin}, \textbf{Xiannian Hu}, \textbf{Guofeng Quan},
\textbf{Qixiang Wang}\\
        \textsuperscript{1}School of Computer Science, Fudan University \\
        \textsuperscript{2}School of Computer Science and Technology, East China Normal University\\
        \{xiaotianzhang21, yzong22\}@m.fudan.edu.cn, \{19307110196, xpqiu\}@fudan.edu.cn \\
        \{zyying, lhe\}@cs.ecnu.edu.cn\\
        qixiangwang@whu.edu.cn\\
}

\maketitle
\renewcommand{\thefootnote}{\fnsymbol{footnote}}
\footnotetext[1]{Equal Contribution.}
\footnotetext[2]{Corresponding Author.}

\renewcommand{\thefootnote}{\arabic{footnote}}

\maketitle
\begin{abstract}
Large Language Models(LLMs) have demonstrated remarkable performance across various natural language processing tasks; however, how to comprehensively and accurately assess their performance becomes an urgent issue to be addressed.
This paper introduces GAOKAO-Bench, an intuitive benchmark that employs questions from the Chinese GAOKAO examination as test samples, including both subjective and objective questions.
To align with human examination methods, we design a method based on zero-shot settings to evaluate the performance of LLMs. With human evaluation, we obtain the converted total score of LLMs, including GPT-4, ChatGPT and ERNIE-Bot.
Our findings reveal that LLMs have achieved competitive scores in Chinese GAOKAO examination, while they exhibit significant performance disparities across various subjects.
We also use LLMs to grade the subjective questions, and find that model scores achieve a moderate level of consistency with human scores.
In conclusion, this research contributes a robust evaluation benchmark for future large language models and offers valuable insights into the advantages and limitations of such models.
\footnote{Resources are available through: \url{https://github.com/OpenLMLab/GAOKAO-Bench}}
\end{abstract}

\section{Introduction}

\begin{figure}[htbp]
    \centering
    \begin{adjustbox}{center}
    \includegraphics[width=1.3\linewidth]{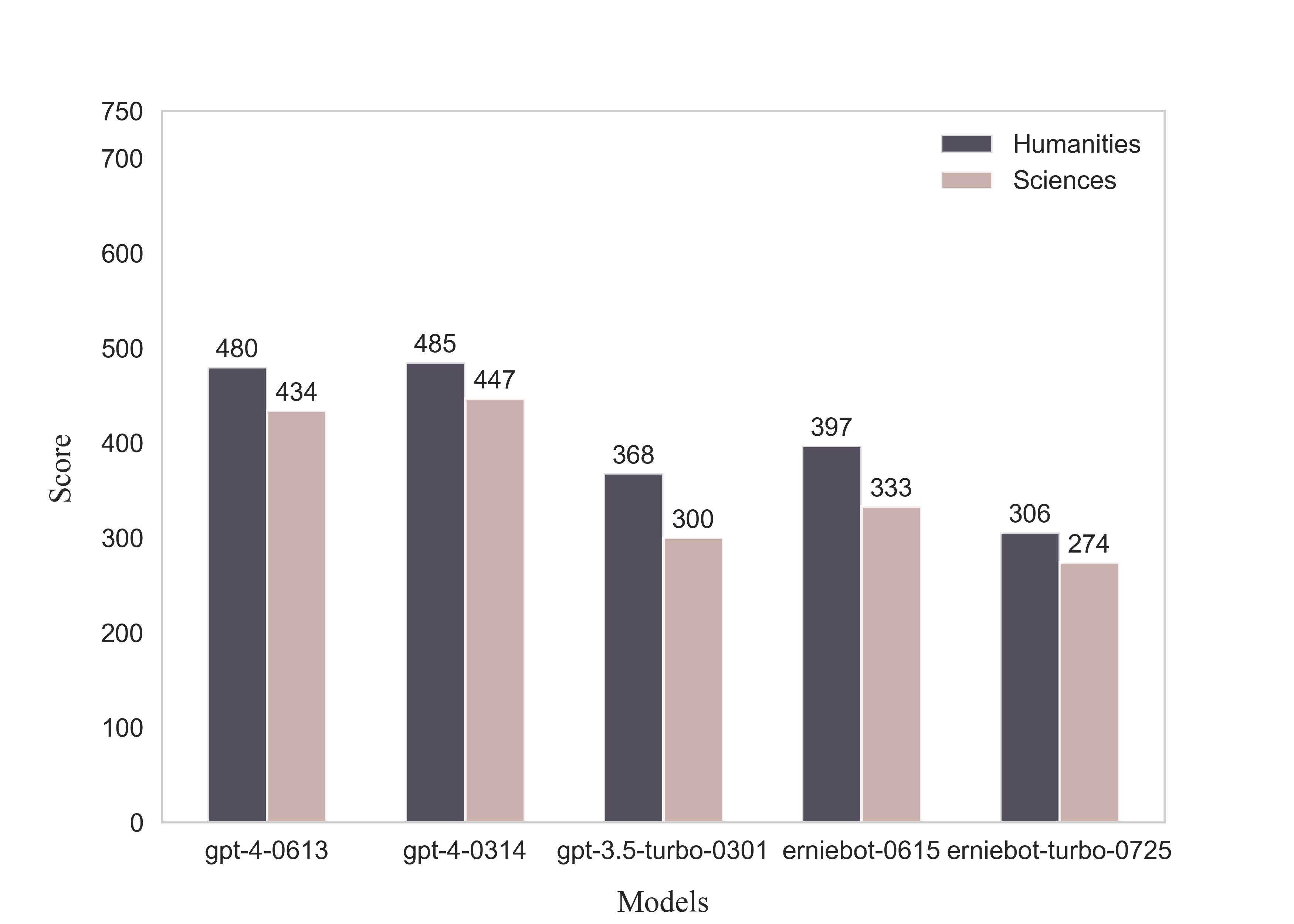}
    \end{adjustbox}
    \caption{\textbf{Converted Total Score of LLMs in GAOKAO-Bench.} The converted total score for subjects in both the sciences and the humanities is 750 points.}
    \label{fig:converted_total_score}
\end{figure}

\begin{figure*}[!ht]
    \centering
    \includegraphics[width=\linewidth]{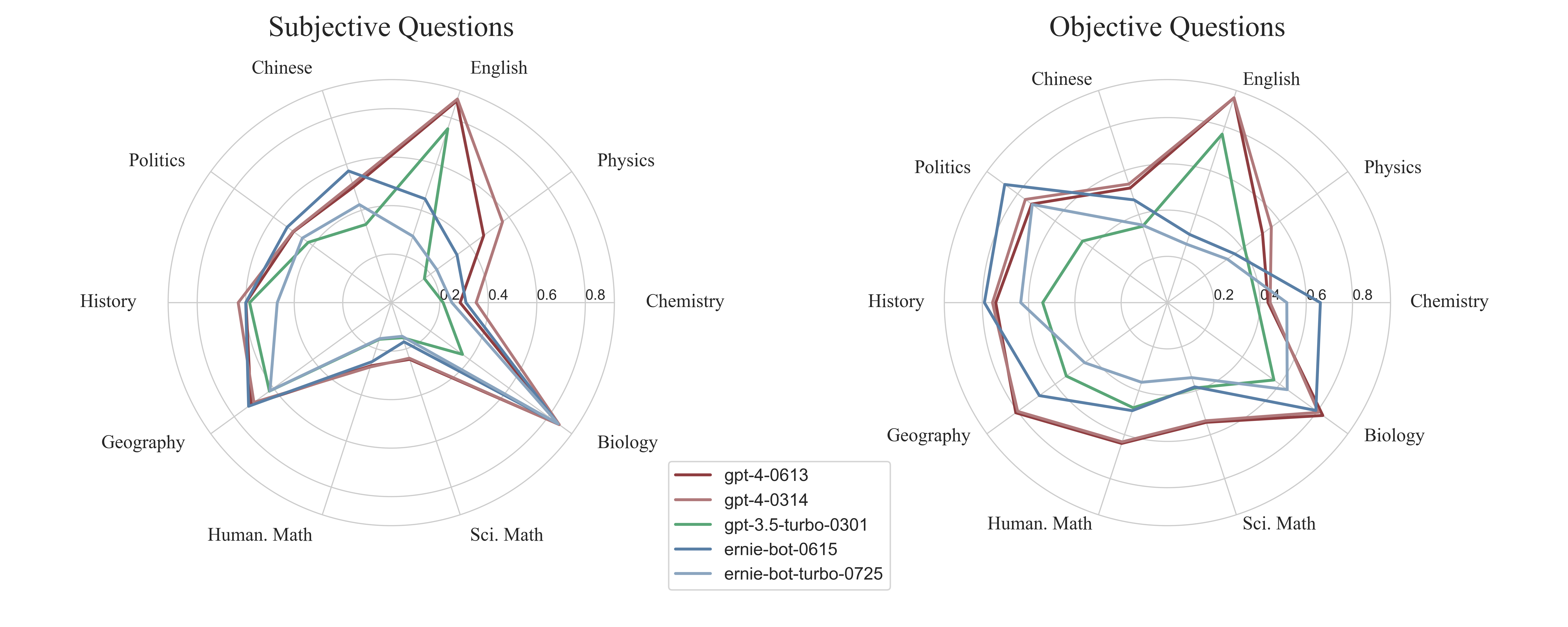}
    \caption{Scoring Rate of LLMs on objective and subjective questions across the subjects.}
    \label{fig:radar map}
\end{figure*}

LLMs have demonstrated great abilities in handling diverse applications. The LLMs (\citealp{brown2020language}; \citealp{ouyang2022training}, \citealp{OpenAI2023GPT4TR}; \citealp{bubeck2023sparks}; \citealp{wei2022chain}) indicate they possess abundant intrinsic knowledge, the ability to follow instructions and reasoning capabilities, which in certain areas are on par with or even surpass human abilities. To better measure the capabilities of LLMs, researchers have proposed more comprehensive and challenging benchmarks. MMLU \citep{hendrycks2021measuring} aims to measure a text model’s multitask accuracy, covering 57 tasks such as elementary mathematics, US history, computer science, and more. BIG-Bench \citep{srivastava2022beyond} introduces a comprehensive evaluation framework encompassing more than 204 subtasks, including linguistics, child development, among others. AGIEval \citep{zhong2023agieval} evaluates the performance of LLMs in the context of human-centric standardized examinations and contains both Chinese and English tasks. \citet{huang2023ceval} propose C-Eval, a comprehensive Chinese evaluation suite covering four difficulty levels. However, the benchmark mentioned above only consists of objective questions and lacks subjective questions that are more closely related to generative abilities. Besides, due to the absence of real-world test samples, individuals often underestimate the complexity of these tasks and the abilities of the models, particularly in the context of the rapid development of LLMs. Consequently, there is a need for an intuitive and practical evaluation method. \\
We propose using the Chinese College Entrance Examination (GAOKAO) questions. These questions include computational, reasoning, knowledge assessment and writing tasks \cite{tan2021gcrc}. Previous benchmarks based on the GAOKAO mainly focus on English \cite{yuan2022restructured}, especially English Reading and Comprehension Questions \cite{zhang2022downstream}. To this end, we introduce the GAOKAO-Benchmark (GAOKAO-Bench), a benchmark specifically tailored to LLMs evaluation that covers the GAOKAO questions from 2010 to 2022. The GAOKAO-Bench consists of 9 subjects with 1781 objective questions and 1030 subjective questions. The question types include single-choice, cloze, correction, open-ended questions, and more. \\
We conduct experiments on some currently best-performing LLMs. To more accurately measure their generative capabilities, we use human scoring evaluation to judge subjective questions. The results in Figure\ref{fig:converted_total_score} show that LLMs have achieved competitive scores in the GAOKAO. Meanwhile, we find that all of the LLMs exhibit obvious signs of subject bias, which informs the future development of LLMs. \\
Due to the high cost of human evaluation, we provide human-annotated marking criteria of subjective questions. And we use LLM as a judge to evaluate LLMs on subjective questions. The results indicate that equipped with the detailed marking criteria, LLMs exhibit high consistency with human teachers, making the large-scale assessment of subjective questions feasible.

\renewcommand{\arraystretch}{1.2}
\begin{table*}[ht!]
\centering
\setcellgapes{5pt} 
\makegapedcells 
\resizebox{\textwidth}{!}{
\begin{tabular}{l|c|cccccccccc} 
\hline
\makecell[l]{\textbf{Models}} & \textbf{Overall} & \textbf{Chinese} & \textbf{Eng.} & \makecell[c]{\textbf{Sci.}\\\textbf{Math}} & \makecell[c]{\textbf{Hum.}\\\textbf{Math}} & \textbf{Phys.} & \textbf{Chem.} & \textbf{Biol.} & \textbf{Poli.} & \textbf{Hist.} & \textbf{Geog.}\\ 
\hline

\textbf{LLaMA-7b}    & 21.1\%    & 16.2\% & 20.5\% & \makecell[c]{24.3\%} & \makecell[c]{26.1\%} & 0.0\% & 22.6\% & 22.7\% & 22.2\% & 19.2\% & 24.2\% \\
\textbf{Vicuna-7b}   & 21.0\%     & 12.0\% & 19.6\% & \makecell[c]{23.8\%} & \makecell[c]{23.4\%} & 7.0\% & 27.4\% & 20.0\% & 20.9\% & 23.0\% & 23.2\% \\
\textbf{Baichuan2-7b-Base}   & 27.2\%  & 16.2\% & 21.2\% & \makecell[c]{24.8\%} & \makecell[c]{24.8\%} & 0.0\% & 23.4\% & 24.0\% & 55.3\% & 32.1\% & 24.2\% \\
\textbf{Baichuan2-7b-Chat}   & 40.5\%  & 31.7\% & 33.0\% & \makecell[c]{26.6\%} & \makecell[c]{28.4\%} & 18.0\% & 26.6\% & 48.0\% & 69.7\% & 57.8\% & 49.5\% \\
\textbf{Baichuan2-13b-Chat}   & 43.9\%  & 26.9\% & 34.7\% & \makecell[c]{23.8\%} & \makecell[c]{31.7\%} & 25.0\% & 40.3\% & 53.3\% & 75.3\% & 59.9\% & 61.1\% \\
\textbf{ChatGLM-6b}  & 30.8\%   & 18.6\% & 17.0\% & \makecell[c]{25.2\%} & \makecell[c]{25.7\%} & 12.5\% & 30.6\% & 24.7\% & 54.1\% & 59.9\% & 25.3\% \\
\textbf{ChatGLM2-6b} & 42.7\%    & 31.1\% & 30.6\% & \makecell[c]{29.0\%} & \makecell[c]{35.8\%} & 24.2\% & 46.0\% & 71.3\% & 55.0\% & 59.2\% & 41.1\% \\

\hline

\textbf{GPT-4-0613} & 71.6\%    & 52.1\% & {\textbf{93.2}}\% & \makecell[c]{\textbf{54.5}\%} & \makecell[c]{\textbf{64.0}\%} & 50.8\% & 43.6\% & \textbf{83.0}\% & 72.5\% & 74.2\% & \textbf{81.1}\% \\
\textbf{GPT-4-0314} &\textbf{72.2}\%     & \textbf{53.9}\% & 93.1\% & \makecell[c]{53.7\%} & \makecell[c]{63.3\%} & \textbf{55.5}\% & 44.4\% & 80.7\% & 75.9\% & 75.6\% & 80.0\% \\
\textbf{GPT-3.5-turbo-0301} &53.2\%  & 34.7\% & 76.6\% & \makecell[c]{38.8\%} & \makecell[c]{47.8\%} & 41.1\% & 38.7\% & 56.9\% & 45.3\% & 53.9\% & 54.0\% \\
\textbf{ERNIE-Bot-0615}   &56.6\%     & 46.7\% & 31.0\% & \makecell[c]{38.3\%} & \makecell[c]{49.1\%} & 35.9\% & \textbf{66.1}\% & 79.3\% & \textbf{86.9}\% & \textbf{79.1}\% & 68.4\% \\
\textbf{ERNIE-Bot-turbo-0725} &45.6\% & 35.3\% & 26.6\% & \makecell[c]{34.1\%} & \makecell[c]{36.2\%} & 32.0\% & 51.6\% & 64.0\% & 72.2\% & 63.4\% & 44.2\% \\ 
\hline
\end{tabular}
}
\caption{\label{tab:scoring_rate_obj}\textbf{Scoring Rate of Objective Questions.} Models above the line are open-source LLMs; models below the line are closed-source LLMs. }
\label{your-table-label}
\end{table*}
\renewcommand{\arraystretch}{1} 

\section{GAOKAO-Bench}

\subsection{Introduction to the GAOKAO}
The Chinese College Entrance Examination, also known as the GAOKAO, is a nationwide examination designed to assess the academic abilities of high school students applying to universities in China. Known as a rigorous and comprehensive examination, the GAOKAO is differentiated into two distinct streams: the sciences and the humanities: the sciences include Chinese, sciences mathematics, English, physics, chemistry and biology; the humanities include Chinese, humanities mathematics, English, politics, history and geography. The examination encompasses a variety of question types that include logical reasoning, computational analysis, knowledge-based quizzes and written expression among other aspects.

\subsection{Dataset Description}
The GAOKAO-Bench established in this paper includes the content of all national exams in the GAOKAO of all subjects from 2010 to 2022, providing an intuitive and human-aligned evaluation benchmark for LLMs.\\
We obtain the questions and transform them from PDF into JSON file format using a combination of automated scripting and manual annotation. Mathematical formulas within the questions were converted into \LaTeX \,format. Appendix \ref{app:data_example} provides an example of a mathematical single-choice question. \\

The questions are divided into subjective and objective categories, depending on whether they require human scoring. In total, we select 2811 questions, including 1030 objective questions and 1781 objective questions. Table \ref{tab:question_distribution} provides a breakdown of the specific types of questions and the corresponding number of questions in each type. Multi-Question Choice refers to a format where a single question is followed by multiple sub-questions and Multi-Choice refers to a format where a single question corresponds to multiple correct answers.

\renewcommand{\arraystretch}{1}
\begin{table*}[ht!]
\centering
\setcellgapes{5pt} 
\makegapedcells 
\resizebox{\textwidth}{!}{
\begin{tabular}{l|c|cccccccccc} 
\hline
\textbf{Models} & \textbf{Overall} & \textbf{Chinese} & \textbf{Eng.} & \makecell[c]{\textbf{Sci.}\\\textbf{Math}} & \makecell[c]{\textbf{Hum.}\\\textbf{Math}} & \textbf{Phys.} & \textbf{Chem.} & \textbf{Biol.} & \textbf{Poli.} & \textbf{Hist.} & \textbf{Geog.}\\ \hline

\textbf{GPT-4-0613} & 50.8\%    & 50.3\% & 87.6\% & \makecell[c]{\textbf{24.6}\%} & \makecell[c]{27.5\%} & 47.1\% & 28.5\% & \textbf{85.6\%} & 49.9\% & 59.9\% & 71.5\% \\

\textbf{GPT-4-0314} &\textbf{51.9\%}     & 51.5\% & \textbf{88.3}\% & \makecell[c]{24.1\%} & \makecell[c]{\textbf{27.9}\%} & \textbf{56.7\%} & \textbf{35.0\%} & \textbf{85.6}\% & 50.0\% & \textbf{63.1\%} & 70.0\% \\

\textbf{GPT-3.5-turbo-0301} &35.8\%  & 33.9\% & 75.4\% & \makecell[c]{15.2\%} & \makecell[c]{15.9\%} & 16.9\% & 21.4\% & 36.3\% & 42.3\% & 58.4\% & 62.1\% \\

\textbf{ERNIE-Bot-0615}   &48.4\%     & \textbf{57.1\%} & 45.0\% & \makecell[c]{17.0\%} & \makecell[c]{25.6\%} & 33.5\% & 30.8\% & 84.9\% & \textbf{53.0\%} & 60.0\% & \textbf{72.7}\% \\

\textbf{ERNIE-Bot-turbo-0725} &39.2\% & 42.5\% & 28.8\% & \makecell[c]{14.6\%} & \makecell[c]{15.6\%} & 23.2\% & 25.0\% & 85.1\% & 45.3\% & 47.0\% & 61.8\% \\ 
\hline
\end{tabular}
}
\caption{\textbf{Scoring Rate of Subjective Questions.} The results are scored by human teachers.}
\label{tab:scoring_rate_subj}
\end{table*}
\renewcommand{\arraystretch}{1} 

\begin{table}[h]
\centering
\resizebox{\linewidth}{!}{
\begin{tabular}{p{2.5cm}l|ccc}
\hline
\multicolumn{2}{l}{\textbf{Question Type}}& \textbf{Number} &\textbf{Percentage} \\ 
\hline
\multirow{4}{*}{\textbf{Objective}}         & Single Choice           & 1418            &   50.5\%              \\
                                            & Multi-Question Choice   & 273             &   9.7\%               \\
                                            & Multi-Choice            & 64              &   2.3\%               \\
                                            & Five out of Seven       & 26              &   0.9\%               \\
\hline
\multirow{3}{*}{\textbf{Subjective}}        & Open-ended Question     & 786             &   28.0\%               \\
                                            & Cloze                   & 218             &   7.8\%               \\ 
                                            & Correction              & 26              &   0.9\%               \\
\hline
\end{tabular}
}
\caption{Distribution of Question Types.}
\label{tab:question_distribution}
\end{table}

\section{Experiments}

\subsection{Methodology}

\textbf{Prompt Design}\quad In order to emulate the format in which humans partake in examinations, we utilize a zero-shot settings strategy \cite{ouyang2022training} and create prompts tailored to different question types. The prompts not only require the model to complete the task, but also explicitly specify the format of the output as we contend that the intrinsic knowledge level of the model and its ability to follow instructions are equally important. The specific prompt examples we use are illustrated in Appendix \ref{app:data_example}.
\\
\\
\textbf{Models}\quad We evaluate several current best-performing LLMs that support both Chinese and English:
\begin{enumerate}
    \item GPT-4: We test on 2 checkpoints: GPT-4-0613 and GPT-4-0314.
    \item ChatGPT: We test on GPT-3.5-turbo-0301 checkpoint. 
    \item ERNIE-Bot: A Chinese LLM published by Baidu. We test on ERNIE-Bot-0615 checkpoint.
    \item ERNIE-Bot-turbo: We test on ERNIE-Bot-turbo-0725 checkpoint.
\end{enumerate}
We set the sampling temperature to 0.3 in order to achieve a balance between stability and diversity.
\\
\\
\textbf{Metric}\quad When evaluating objective and subjective questions separately, we use the scoring rate $R_{i, \text{obj}}$ and $R_{i, \text{subj}}$ for each subject ${i}$. \\
To evaluate the overall performance, we convert the scoring rates of subjective and objective questions into a total score $S_{\text{total}}$. We mimic the subjective question scores as $M_{i, \text{subj}}$ and objective question scores $M_{i, \text{obj}}$ for each subject ${i}$ in the GAOKAO. The converted total score can be formulated as:

\begin{align*}
S_{\text{total, $\mathcal{S}$}} &= \sum_{i \in \mathcal{S}}\left(R_{i, \text{obj}} \cdot M_{i, \text{obj}} + R_{i, \text{subj}} \cdot M_{i, \text{subj}}\right), \\
S_{\text{total, $\mathcal{H}$}} &= \sum_{i \in \mathcal{H}}\left(R_{i, \text{obj}} \cdot M_{i, \text{obj}} + R_{i, \text{subj}} \cdot M_{i, \text{subj}}\right)
\end{align*}\\

where $\mathcal{S}$ stands for the set of the sciences subjects, and $\mathcal{H}$ stands for the set of the humanities subjects. The total scores of sciences and humanities are both 750 points. Detailed total score for each subject is shown in Appendix \ref{app:cvt_score}.

\subsection{Objective Questions}
Each item $i$ in the GAOKAO-Bench comprises the question $q_i$, the standard answer $a_i$, the score $s_i$, the analysis $n_i$. For objective questions, the input includes the question $q_i$ and the LLMs need to output $(r_i, o_i)$, where $r_i$ denotes the corresponding reasoning process and $o_i$ denotes the outcome. Points are awarded only if the outcome $o_i$ is consistent with the standard answer $a_i$. Following the technical report for OpenAI's GPT-4 \cite{OpenAI2023GPT4TR}, we score the objective questions using regular matching. In addition to the LLMs mentioned above, we evaluate several open-source LLMs on GAOKAO-Bench, including LLaMA \cite{touvron2023LLaMA}, Baichuan \cite{yang2023baichuan}and ChatGLM \cite{zeng2023glm130b}.

\subsection{Subjective Questions}
The input and output formats of the subjective questions are similar to those of objective questions. During the grading process, evaluators take into account both the reasoning process $r_i$ and the outcome $o_i$. We assess the subjective questions using human scoring, in order to more precisely evaluate the performance of LLMs. 
Each subjective question is evaluated by two teachers, and the average of these scores was adopted as the final score for that question. 

\subsection{LLM as a Judge}
Due to the high cost of manual evaluation, it is a natural progression to consider the use of LLMs for grading subjective questions. To better align with the teachers, we solicit teachers to provide detailed marking criteria $m_i$, breaking down the answers into specific scoring points for each item $i$. We design prompts in zero-shot settings and utilize GPT-4-turbo (GPT-4-1106-preview) as a judge. For each input $(q_i, a_i, s_i, n_i, m_i, r_i, o_i)$, the LLM need to output $(g_i, f_i)$, where $g_i$ denotes the process of grading and $f_i$ denotes the final score. The sampling temperature is set to 0 to obtain deterministic scores. We calculate the converted total score and Spearman and Kendall-Tau correlations between predicted scores and human scores following \citet{jain2023multidimensional} and \citet{zhong2022unified}.

\subsection{Results}
\textbf{Overall Performance}\quad Figure \ref{fig:converted_total_score} shows the converted total score of LLMs on GAOKAO-Bench. GPT-4 achieves scores exceeding 400 points and ERNIE-Bot surpasses ChatGPT. Every LLM obtains higher scores in humanities than in sciences. In the GAOKAO, the sciences require more advanced logical reasoning and computational steps than the humanities; and the humanities require a greater amount of knowledge than the sciences. The result indicates the reasoning and calculation abilities of LLMs still need further improvement.\\

\textbf{Performance on Objective Questions}\quad Table \ref{tab:scoring_rate_obj} reflects the performance of LLMs on objective questions in different subjects. Open-source models pre-trained on Chinese language data and aligned with human perform better in all subjects. And the performance of the models improves with the increase in their scale. For closed-source LLMs, GPT-4 maintains a lead in the majority of subjects, but ERNIE-Bot performs better in chemistry, politics and history.\\

\textbf{Performance on Subjective Questions}\quad Table \ref{tab:scoring_rate_subj} indicates the human evaluation of subjective questions. GPT-4 obtains the highest scoring rate (51.9\%) and ERNIE-Bot achieves a comparably close level (48.4\%). GPT-4 and ChatGPT exhibit superior performance in English compared to Chinese, whereas ERNIE-Bot and ERNIE-Bot-turbo demonstrate the opposite trend, excelling more in Chinese than in English. \\

\textbf{LLM as a Judge}\quad Table \ref{tab:llms as teachers} shows the results of using GPT-4-turbo to grade subjective questions. The Question-level Spearman and Kendall-Tau correlations show a markedly strong positive correlation between model judging and human scoring.

\renewcommand{\arraystretch}{1}
\begin{table*}[ht!]
\centering
\setcellgapes{5pt} 
\makegapedcells 
\resizebox{0.8\textwidth}{!}{

\begin{tabular}{l|cc|cc|c|c} 
\hline
\multirow{2}{*}{\textbf{Models}} & \multicolumn{2}{c|}{\textbf{Sciences}} & \multicolumn{2}{c|}{\textbf{Humanities}} & \multirow{2}{*}{\textbf{$\rho$}} & \multirow{2}{*}{\textbf{$\tau$}}\\ 
 & \textbf{Human} & \textbf{GPT-4-turbo} & \textbf{Human} & \textbf{GPT-4-turbo} & & \\
 \hline
\textbf{GPT-4-0613} & 434 & 428 & 480 & 523 & 0.854 & 0.710 \\

\textbf{GPT-4-0314} & 447 & 440 & 485 & 535 & 0.852 & 0.709\\

\textbf{GPT-3.5-turbo-0301} & 300 & 314 & 368 & 408 & 0.845 & 0.710\\

\textbf{ERNIE-Bot-0615} & 333 & 348 & 397 & 436 & 0.864 & 0.740  \\

\textbf{ERNIE-Bot-turbo-0725} & 274 & 262 & 306 & 328 & 0.825 & 0.685 \\ 

\hline
\end{tabular}
}

\caption{Converted total score, Question-level Spearman and Kendall-Tau correlations of Human and GPT-4-turbo grading methods.}
\label{tab:llms as teachers}
\end{table*}
\renewcommand{\arraystretch}{1} 

\section{Analysis}

\subsection{Difference in Subjects}
We analyze the scoring rate of subjective questions and objective questions in different subjects of LLMs, and find that there are large differences in the ability of the model in different subjects both in objective questions and subjective questions.\\
GPT-4 excels in English, biology and geography with scoring rates greater than 70\% both in subjective and objective questions. However, they demonstrate poor performance in mathematics and physics with scoring rates less than 40\%. ERNIE-Bot performs better in biology, history, politics in subjective questions with scoring rates greater than 60\%, but the scoring rate of mathematics is less than 30\%. \\
We posit that the substantial disparities across subjects can be attributed to two primary factors: firstly, the distinct competencies evaluated by each subject, for instance, language comprehension and summarization abilities in Chinese and English, and logical reasoning and computational skills in mathematics and physics; secondly, aspects related to the training of the model, including the richness of the pre-training corpus and the inclinations towards human alignment.

\subsection{Difference between Sujective and Objective Questions}

For a given subject, the scoring rate of subjective questions is generally lower than that of objective questions. For example, the scoring rate of subjective mathematics questions of GPT-4 is significantly lower than that on subjective mathematics questions. We hypothesize that subjective mathematics questions distinctly require the application of correct formulas, as well as more extensive computational and reasoning steps, which poses a significant challenge for LLMs. And compared to objective questions, the subjective questions of humanities necessitate students' mastery of more precise knowledge points, as well as their abilities in induction, summarization and categorical organization.

\begin{figure}[!htb]
    \centering
    \includegraphics[width=\linewidth,scale=1.00]{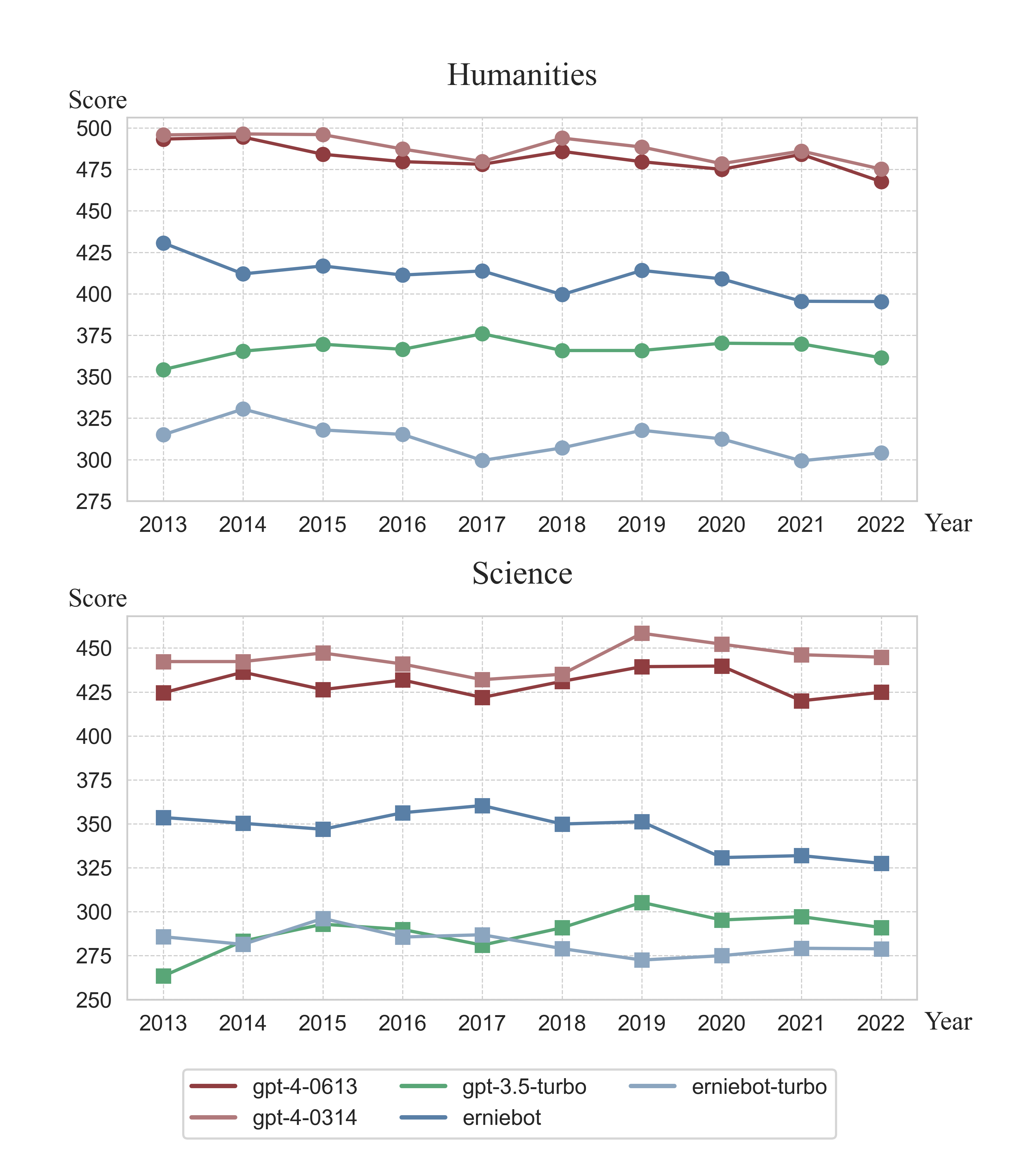}
    \caption{The Annual Trends of LLMs on GAOKAO-Bench.}
    \label{fig:line_chart}
\end{figure}

\subsection{Stable Annual Trends on the GAOKAO}
We categorize the examination questions based on their respective years and compute the model's converted total scores from 2013 to 2022 in Figure \ref{fig:line_chart}. We observe that the converted total score of LLMs are stable across the last decade. It indicates a relative stability in the difficulty level of the GAOKAO questions. 

\subsection{LLM as a Judge}
We observe that the converted total score of sciences is much closer to human scoring than that of humanities. The deviation in scores for science subjects is less than 2\% of the total score, and for humanities subjects, the deviation is around 5\% of the total score. We posit that the answers and scoring criteria in the sciences are relatively explicit, whereas in the humanities, scoring depends on the alignment of semantics with designated points. This necessitates a fine-grained semantic understanding by the models, presenting a significant challenge for LLMs.

\section{Avoid Benchmark Leakage}
Benchmark leakage means the data related to evaluation sets is occasionally used for model training \citep{zhou2023dont}. And it is plausible that the GAOKAO questions may be included in the training corpus of LLMs. The zero-shot settings and human evaluation used in this paper can alleviate the unfair phenomenon. Given that the GAOKAO is conducted annually in June, we plan to incorporate each year's new GAOKAO questions into the GAOKAO-Bench as a supplement, aiming to mitigate the issue of dataset leakage in evaluations.  We have released the GAOKAO-Bench-2023
\footnote{Resources are available through: \url{https://github.com/OpenLMLab/GAOKAO-Bench-2023}} 
which includes the objective questions in the 2023 GAOKAO. And we compare the scoring rate of objective questions in GAOKAO-Bench and GAOKAO-Bench-2023 in Table \ref{tab:GAOKAO-Bench-2023}. We contend that these variations are within the normal range of difficulty fluctuations.

\renewcommand{\arraystretch}{1.2}
\begin{table}[h]
\centering
\resizebox{\linewidth}{!}{
\begin{tabular}{l|ccc}
\hline
\textbf{Models}& \textbf{GAOKAO-Bench} &\textbf{GAOKAO-Bench-2023} &\textbf{$\Delta$} \\ 
\hline
\textbf{ChatGLM-6b}         & 30.8\%    & 24.1\%    & -6.7\%    \\
\textbf{ChatGLM2-6b}        & 42.7\%    &36.9\%     & -5.8\%    \\
\textbf{Baichuan2-7b-chat}  &40.5\%     &37.9\%     &-2.6\%     \\
\textbf{Baichuan2-13b-chat} &43.9\%     &41.3\%     &-2.6\%     \\
\textbf{GPT-4-0613}         & 71.6\%    & 71.0\%    &-0.6\%     \\
\textbf{GPT-4-0314}         & 72.2\%    & 69.8\%    & -2.4\%    \\
\hline
\end{tabular}
}
\caption{\textbf{Scoring Rate of Objective Questions on GAOKAO-Bench-2023.} The GAOKAO-Bench covers questions from 2010 to 2022.}
\label{tab:GAOKAO-Bench-2023}
\end{table}
\renewcommand{\arraystretch}{1} 

\section{Ablation Study}
We investigate the impact of manually annotated marking criteria on the accuracy of the LLM's grading of subjective questions. We use the GPT-4-turbo to evaluate the performance of GPT-4, ChatGPT and ERNIE-Bot-turbo with or without marking criteria. Tabel \ref{tab:w/o marking criterion} indicates that provided with marking criteria, LLMs can better align with human preferences.

\renewcommand{\arraystretch}{1.2}
\begin{table}[h]
\centering
\resizebox{\linewidth}{!}{
\begin{tabular}{lc|ccc}
\hline
\textbf{Methods}& &\textbf{GPT-4-0613} &\textbf{GPT-3.5-turbo-0301} &\textbf{ERNIE-Bot-turbo-0725}\\ 
\hline
\multirow{2}{*}{\textbf{w marking criterion}}   &  $\rho$     & 0.854       &   0.845    &   0.825\\
                                                &  $\tau$     & 0.710       &   0.710    &   0.685\\
                                                \hline
\multirow{2}{*}{\textbf{w/o marking criterion}} &  $\rho$     & 0.820       &   0.820    &   0.803\\
                                                &  $\tau$     & 0.659       &   0.674    &   0.654\\
\hline

\end{tabular}
}
\caption{Spearman and Kendall-Tau Correlations of LLM grading and human judgement.}
\label{tab:w/o marking criterion}
\end{table}
\renewcommand{\arraystretch}{1} 

\section{Related Work}
\textbf{Benchmark for LLMs}\quad The flourishing development of LLMs has also raised higher demands for benchmarks. Benchmarks for traditional tasks in NLP, such as GLUE \citep{wang-etal-2018-glue} for natural language understanding, SQuAD \citep{rajpurkar2016squad} for reading comprehension, cannot measure the comprehensive capabilities of LLMs. Consequently, researchers have proposed new benchmarks to evaluate the advanced abilities of LLMs. MMLU \citep{hendrycks2021measuring} provides a multi-task test across a diverse set of subjects. BIG-Bench \citep{srivastava2022beyond} covers a diverse range of topics and languages, including auto debugging, know unknowns, logical deduction. HELM \citep{liang2023holistic} taxonomies the design space of language model evaluation into scenarios and metrics. In the field of Chinese language benchmarks, C-Eval \citep{huang2023ceval} selects multiple-choice questions across four difficulty levels: middle school, high school, college, and professional. AGIEval \citep{zhong2023agieval} assesses LLMs in the context of human-centric standardized exams. CMMLU \citep{li2023cmmlu} includes subjects that may not typically appear in standard exams but are relevant to people’s daily life, such as Chinese food culture, Chinese driving rule.\\
\\
\textbf{Human evaluation for LLMs}\quad Compared to automatic evaluation, human evaluation is more aligned with real-world application scenarios and can offer more comprehensive and precise feedback \citep{chang2023survey}. Chatbot Arena \citep{zheng2023judging} provides a platform to assess and compare diverse chatbot models through user engagement and voting. \citet{ziems2023large} adopts human scoring evaluation on generation tasks. \citet{liang2023holistic} conduct human evaluations on 6 LLMs on summarization and disinformation scenarios.

\section{Limitations}
While we evaluate and analyze the performance of LLMs on GAOKAO-Bench, there are some limitations in this work. Firstly, due to the constraints in time and resources, this paper does not delve into the detailed analysis of the errors made by LLMs on the GAOKAO-Bench, such as model hallucinations and reasoning mistakes. Secondly, due to the rapid developments of LLMs and high cost of human evaluation, we are unable to conduct experiments on every model using human scoring. We hope to enhance the evaluation and analysis of the models' reasoning process and utilize LLMs as a replacement for human scoring in future work.\\

\section{Conclusion}

In this paper, we introduce the GAOKAO-Bench dataset, which serves as an evaluation standard for large language models. The dataset includes Chinese College Entrance Examination questions from 2010 to 2022, covering various subjects and question types, with an overall high level of difficulty. By testing large language models on the GAOKAO-Bench, we can analyze the gap and advantages of these models compared to humans in a reasonable and intuitive manner.

In addition, we evaluate the ability of large language models to answer Chinese College Entrance Examination questions using zero-shot prediction approach and human evaluation. Our results show that the models perform well on knowledge-based questions, but struggle with certain types of logical reasoning and mathematical problems, as well as with reading comprehension of longer texts in Chinese. 

We also use the LLMs to evaluate subjective questions, which is called LLM-as-a-Judge. We observe that equipped with human-annotated marking criteria, the LLM evaluation is consistent to human preference. 

These findings suggest that large language models have potential applications in education and language assessment, but there is still room for improvement in certain areas. Future work could focus on developing approaches to enhance the model's performance on longer text reading comprehension tasks, logical reasoning and calculation problems.

\section*{Author Contributions}
Qixiang Wang contributed to data collection and curation for GAOKAO-Bench.

\bibliography{custom}

\begin{thebibliography}{25}
\expandafter\ifx\csname natexlab\endcsname\relax\def\natexlab#1{#1}\fi

\bibitem[{Brown et~al.(2020)Brown, Mann, Ryder, Subbiah, Kaplan, Dhariwal, Neelakantan, Shyam, Sastry, Askell et~al.}]{brown2020language}
Tom Brown, Benjamin Mann, Nick Ryder, Melanie Subbiah, Jared~D Kaplan, Prafulla Dhariwal, Arvind Neelakantan, Pranav Shyam, Girish Sastry, Amanda Askell, et~al. 2020.
\newblock Language models are few-shot learners.
\newblock \emph{Advances in neural information processing systems}, 33:1877--1901.

\bibitem[{Bubeck et~al.(2023)Bubeck, Chandrasekaran, Eldan, Gehrke, Horvitz, Kamar, Lee, Lee, Li, Lundberg, Nori, Palangi, Ribeiro, and Zhang}]{bubeck2023sparks}
Sébastien Bubeck, Varun Chandrasekaran, Ronen Eldan, Johannes Gehrke, Eric Horvitz, Ece Kamar, Peter Lee, Yin~Tat Lee, Yuanzhi Li, Scott Lundberg, Harsha Nori, Hamid Palangi, Marco~Tulio Ribeiro, and Yi~Zhang. 2023.
\newblock \href {http://arxiv.org/abs/2303.12712} {Sparks of artificial general intelligence: Early experiments with gpt-4}.

\bibitem[{Chang et~al.(2023)Chang, Wang, Wang, Wu, Yang, Zhu, Chen, Yi, Wang, Wang, Ye, Zhang, Chang, Yu, Yang, and Xie}]{chang2023survey}
Yupeng Chang, Xu~Wang, Jindong Wang, Yuan Wu, Linyi Yang, Kaijie Zhu, Hao Chen, Xiaoyuan Yi, Cunxiang Wang, Yidong Wang, Wei Ye, Yue Zhang, Yi~Chang, Philip~S. Yu, Qiang Yang, and Xing Xie. 2023.
\newblock \href {http://arxiv.org/abs/2307.03109} {A survey on evaluation of large language models}.

\bibitem[{Hendrycks et~al.(2021)Hendrycks, Burns, Basart, Zou, Mazeika, Song, and Steinhardt}]{hendrycks2021measuring}
Dan Hendrycks, Collin Burns, Steven Basart, Andy Zou, Mantas Mazeika, Dawn Song, and Jacob Steinhardt. 2021.
\newblock \href {https://openreview.net/forum?id=d7KBjmI3GmQ} {Measuring massive multitask language understanding}.
\newblock In \emph{International Conference on Learning Representations}.

\bibitem[{Huang et~al.(2023)Huang, Bai, Zhu, Zhang, Zhang, Su, Liu, Lv, Zhang, Lei, Fu, Sun, and He}]{huang2023ceval}
Yuzhen Huang, Yuzhuo Bai, Zhihao Zhu, Junlei Zhang, Jinghan Zhang, Tangjun Su, Junteng Liu, Chuancheng Lv, Yikai Zhang, Jiayi Lei, Yao Fu, Maosong Sun, and Junxian He. 2023.
\newblock \href {http://arxiv.org/abs/2305.08322} {C-eval: A multi-level multi-discipline chinese evaluation suite for foundation models}.

\bibitem[{Jain et~al.(2023)Jain, Keshava, Sathyendra, Fernandes, Liu, Neubig, and Zhou}]{jain2023multidimensional}
Sameer Jain, Vaishakh Keshava, Swarnashree~Mysore Sathyendra, Patrick Fernandes, Pengfei Liu, Graham Neubig, and Chunting Zhou. 2023.
\newblock \href {http://arxiv.org/abs/2306.01200} {Multi-dimensional evaluation of text summarization with in-context learning}.

\bibitem[{Li et~al.(2023)Li, Zhang, Koto, Yang, Zhao, Gong, Duan, and Baldwin}]{li2023cmmlu}
Haonan Li, Yixuan Zhang, Fajri Koto, Yifei Yang, Hai Zhao, Yeyun Gong, Nan Duan, and Timothy Baldwin. 2023.
\newblock \href {http://arxiv.org/abs/2306.09212} {Cmmlu: Measuring massive multitask language understanding in chinese}.

\bibitem[{Liang et~al.(2023)Liang, Bommasani, Lee, Tsipras, Soylu, Yasunaga, Zhang, Narayanan, Wu, Kumar, Newman, Yuan, Yan, Zhang, Cosgrove, Manning, Ré, Acosta-Navas, Hudson, Zelikman, Durmus, Ladhak, Rong, Ren, Yao, Wang, Santhanam, Orr, Zheng, Yuksekgonul, Suzgun, Kim, Guha, Chatterji, Khattab, Henderson, Huang, Chi, Xie, Santurkar, Ganguli, Hashimoto, Icard, Zhang, Chaudhary, Wang, Li, Mai, Zhang, and Koreeda}]{liang2023holistic}
Percy Liang, Rishi Bommasani, Tony Lee, Dimitris Tsipras, Dilara Soylu, Michihiro Yasunaga, Yian Zhang, Deepak Narayanan, Yuhuai Wu, Ananya Kumar, Benjamin Newman, Binhang Yuan, Bobby Yan, Ce~Zhang, Christian Cosgrove, Christopher~D. Manning, Christopher Ré, Diana Acosta-Navas, Drew~A. Hudson, Eric Zelikman, Esin Durmus, Faisal Ladhak, Frieda Rong, Hongyu Ren, Huaxiu Yao, Jue Wang, Keshav Santhanam, Laurel Orr, Lucia Zheng, Mert Yuksekgonul, Mirac Suzgun, Nathan Kim, Neel Guha, Niladri Chatterji, Omar Khattab, Peter Henderson, Qian Huang, Ryan Chi, Sang~Michael Xie, Shibani Santurkar, Surya Ganguli, Tatsunori Hashimoto, Thomas Icard, Tianyi Zhang, Vishrav Chaudhary, William Wang, Xuechen Li, Yifan Mai, Yuhui Zhang, and Yuta Koreeda. 2023.
\newblock \href {http://arxiv.org/abs/2211.09110} {Holistic evaluation of language models}.

\bibitem[{OpenAI(2023)}]{OpenAI2023GPT4TR}
OpenAI. 2023.
\newblock Gpt-4 technical report.
\newblock \emph{ArXiv}, abs/2303.08774.

\bibitem[{Ouyang et~al.(2022)Ouyang, Wu, Jiang, Almeida, Wainwright, Mishkin, Zhang, Agarwal, Slama, Ray et~al.}]{ouyang2022training}
Long Ouyang, Jeffrey Wu, Xu~Jiang, Diogo Almeida, Carroll Wainwright, Pamela Mishkin, Chong Zhang, Sandhini Agarwal, Katarina Slama, Alex Ray, et~al. 2022.
\newblock Training language models to follow instructions with human feedback.
\newblock \emph{Advances in Neural Information Processing Systems}, 35:27730--27744.

\bibitem[{Rajpurkar et~al.(2016)Rajpurkar, Zhang, Lopyrev, and Liang}]{rajpurkar2016squad}
Pranav Rajpurkar, Jian Zhang, Konstantin Lopyrev, and Percy Liang. 2016.
\newblock \href {http://arxiv.org/abs/1606.05250} {Squad: 100,000+ questions for machine comprehension of text}.

\bibitem[{Srivastava et~al.(2022)Srivastava, Rastogi, Rao, Shoeb, Abid, Fisch, Brown, Santoro, Gupta, Garriga-Alonso et~al.}]{srivastava2022beyond}
Aarohi Srivastava, Abhinav Rastogi, Abhishek Rao, Abu Awal~Md Shoeb, Abubakar Abid, Adam Fisch, Adam~R Brown, Adam Santoro, Aditya Gupta, Adri{\`a} Garriga-Alonso, et~al. 2022.
\newblock Beyond the imitation game: Quantifying and extrapolating the capabilities of language models.
\newblock \emph{arXiv preprint arXiv:2206.04615}.

\bibitem[{Tan et~al.(2021)Tan, Wang, Ji, Li, Li, Hu, Zhao, and Han}]{tan2021gcrc}
Hongye Tan, Xiaoyue Wang, Yu~Ji, Ru~Li, Xiaoli Li, Zhiwei Hu, Yunxiao Zhao, and Xiaoqi Han. 2021.
\newblock Gcrc: A new challenging mrc dataset from gaokao chinese for explainable evaluation.
\newblock In \emph{Findings of the Association for Computational Linguistics: ACL-IJCNLP 2021}, pages 1319--1330.

\bibitem[{Touvron et~al.(2023)Touvron, Lavril, Izacard, Martinet, Lachaux, Lacroix, Rozière, Goyal, Hambro, Azhar, Rodriguez, Joulin, Grave, and Lample}]{touvron2023LLaMA}
Hugo Touvron, Thibaut Lavril, Gautier Izacard, Xavier Martinet, Marie-Anne Lachaux, Timothée Lacroix, Baptiste Rozière, Naman Goyal, Eric Hambro, Faisal Azhar, Aurelien Rodriguez, Armand Joulin, Edouard Grave, and Guillaume Lample. 2023.
\newblock \href {http://arxiv.org/abs/2302.13971} {Llama: Open and efficient foundation language models}.

\bibitem[{Wang et~al.(2018)Wang, Singh, Michael, Hill, Levy, and Bowman}]{wang-etal-2018-glue}
Alex Wang, Amanpreet Singh, Julian Michael, Felix Hill, Omer Levy, and Samuel Bowman. 2018.
\newblock \href {https://doi.org/10.18653/v1/W18-5446} {{GLUE}: A multi-task benchmark and analysis platform for natural language understanding}.
\newblock In \emph{Proceedings of the 2018 {EMNLP} Workshop {B}lackbox{NLP}: Analyzing and Interpreting Neural Networks for {NLP}}, pages 353--355, Brussels, Belgium. Association for Computational Linguistics.

\bibitem[{Wei et~al.(2022)Wei, Wang, Schuurmans, Bosma, Chi, Le, and Zhou}]{wei2022chain}
Jason Wei, Xuezhi Wang, Dale Schuurmans, Maarten Bosma, Ed~Chi, Quoc Le, and Denny Zhou. 2022.
\newblock Chain of thought prompting elicits reasoning in large language models.
\newblock \emph{arXiv preprint arXiv:2201.11903}.

\bibitem[{Yang et~al.(2023)Yang, Xiao, Wang, Zhang, Bian, Yin, Lv, Pan, Wang, Yan, Yang, Deng, Wang, Liu, Ai, Dong, Zhao, Xu, Sun, Zhang, Liu, Ji, Xie, Dai, Fang, Su, Song, Liu, Ru, Ma, Wang, Liu, Lin, Nie, Guo, Sun, Zhang, Li, Li, Cheng, Chen, Zeng, Wang, Chen, Men, Yu, Pan, Shen, Wang, Li, Jiang, Gao, Zhang, Zhou, and Wu}]{yang2023baichuan}
Aiyuan Yang, Bin Xiao, Bingning Wang, Borong Zhang, Ce~Bian, Chao Yin, Chenxu Lv, Da~Pan, Dian Wang, Dong Yan, Fan Yang, Fei Deng, Feng Wang, Feng Liu, Guangwei Ai, Guosheng Dong, Haizhou Zhao, Hang Xu, Haoze Sun, Hongda Zhang, Hui Liu, Jiaming Ji, Jian Xie, JunTao Dai, Kun Fang, Lei Su, Liang Song, Lifeng Liu, Liyun Ru, Luyao Ma, Mang Wang, Mickel Liu, MingAn Lin, Nuolan Nie, Peidong Guo, Ruiyang Sun, Tao Zhang, Tianpeng Li, Tianyu Li, Wei Cheng, Weipeng Chen, Xiangrong Zeng, Xiaochuan Wang, Xiaoxi Chen, Xin Men, Xin Yu, Xuehai Pan, Yanjun Shen, Yiding Wang, Yiyu Li, Youxin Jiang, Yuchen Gao, Yupeng Zhang, Zenan Zhou, and Zhiying Wu. 2023.
\newblock \href {http://arxiv.org/abs/2309.10305} {Baichuan 2: Open large-scale language models}.

\bibitem[{Yuan and Liu(2022)}]{yuan2022restructured}
Weizhe Yuan and Pengfei Liu. 2022.
\newblock restructured pre-training.
\newblock \emph{arXiv preprint arXiv:2206.11147}.

\bibitem[{Zeng et~al.(2023)Zeng, Liu, Du, Wang, Lai, Ding, Yang, Xu, Zheng, Xia, Tam, Ma, Xue, Zhai, Chen, Zhang, Dong, and Tang}]{zeng2023glm130b}
Aohan Zeng, Xiao Liu, Zhengxiao Du, Zihan Wang, Hanyu Lai, Ming Ding, Zhuoyi Yang, Yifan Xu, Wendi Zheng, Xiao Xia, Weng~Lam Tam, Zixuan Ma, Yufei Xue, Jidong Zhai, Wenguang Chen, Peng Zhang, Yuxiao Dong, and Jie Tang. 2023.
\newblock \href {http://arxiv.org/abs/2210.02414} {Glm-130b: An open bilingual pre-trained model}.

\bibitem[{Zhang et~al.(2022)Zhang, Zhang, and Wang}]{zhang2022downstream}
Cheng Zhang, Hao Zhang, and Jie Wang. 2022.
\newblock \href {http://arxiv.org/abs/2205.07387} {Downstream transformer generation of question-answer pairs with preprocessing and postprocessing pipelines}.

\bibitem[{Zheng et~al.(2023)Zheng, Chiang, Sheng, Zhuang, Wu, Zhuang, Lin, Li, Li, Xing, Zhang, Gonzalez, and Stoica}]{zheng2023judging}
Lianmin Zheng, Wei-Lin Chiang, Ying Sheng, Siyuan Zhuang, Zhanghao Wu, Yonghao Zhuang, Zi~Lin, Zhuohan Li, Dacheng Li, Eric.~P Xing, Hao Zhang, Joseph~E. Gonzalez, and Ion Stoica. 2023.
\newblock \href {http://arxiv.org/abs/2306.05685} {Judging llm-as-a-judge with mt-bench and chatbot arena}.

\bibitem[{Zhong et~al.(2022)Zhong, Liu, Yin, Mao, Jiao, Liu, Zhu, Ji, and Han}]{zhong2022unified}
Ming Zhong, Yang Liu, Da~Yin, Yuning Mao, Yizhu Jiao, Pengfei Liu, Chenguang Zhu, Heng Ji, and Jiawei Han. 2022.
\newblock \href {http://arxiv.org/abs/2210.07197} {Towards a unified multi-dimensional evaluator for text generation}.

\bibitem[{Zhong et~al.(2023)Zhong, Cui, Guo, Liang, Lu, Wang, Saied, Chen, and Duan}]{zhong2023agieval}
Wanjun Zhong, Ruixiang Cui, Yiduo Guo, Yaobo Liang, Shuai Lu, Yanlin Wang, Amin Saied, Weizhu Chen, and Nan Duan. 2023.
\newblock \href {http://arxiv.org/abs/2304.06364} {Agieval: A human-centric benchmark for evaluating foundation models}.

\bibitem[{Zhou et~al.(2023)Zhou, Zhu, Chen, Chen, Zhao, Chen, Lin, Wen, and Han}]{zhou2023dont}
Kun Zhou, Yutao Zhu, Zhipeng Chen, Wentong Chen, Wayne~Xin Zhao, Xu~Chen, Yankai Lin, Ji-Rong Wen, and Jiawei Han. 2023.
\newblock \href {http://arxiv.org/abs/2311.01964} {Don't make your llm an evaluation benchmark cheater}.

\bibitem[{Ziems et~al.(2023)Ziems, Held, Shaikh, Chen, Zhang, and Yang}]{ziems2023large}
Caleb Ziems, William Held, Omar Shaikh, Jiaao Chen, Zhehao Zhang, and Diyi Yang. 2023.
\newblock \href {http://arxiv.org/abs/2305.03514} {Can large language models transform computational social science?}

\end{thebibliography}
\begin{CJK*}{UTF8}{gbsn}

\appendix

\section{Examples}
\subsection{Data Example}
\label{app:data_example}
An example of the single-choice questions in sciences mathematics is shown below.
The English is added by the author for readers' understanding.

\begin{tcolorbox}[breakable,colback=black!5!white,colframe=black!75!black]
\begin{itemize}
    \item \textbf{Question}
\end{itemize}
若 $z=-1+\sqrt{3} \mathrm{i}$, 则 $\frac{z}{z \bar{z}-1}=(\quad)$\\
A. $-1+\sqrt{3} \mathrm{i}$\\
B. $-1-\sqrt{3} i$\\
C. $-\frac{1}{3}+\frac{\sqrt{3}}{3} \mathrm{i}$\\
D. $-\frac{1}{3}-\frac{\sqrt{3}}{3} i$\\
\\
If $z=-1+\sqrt{3} \mathrm{i}$, then $\frac{z}{z \bar{z}-1}=(\quad)$\\
A. $-1+\sqrt{3} \mathrm{i}$\\
B. $-1-\sqrt{3} i$\\
C. $-\frac{1}{3}+\frac{\sqrt{3}}{3} \mathrm{i}$\\
D. $-\frac{1}{3}-\frac{\sqrt{3}}{3} i$\\

\begin{itemize}
    \item \textbf{Answer}
\end{itemize}
C
\\
\begin{itemize}
    \item \textbf{Explanation}
\end{itemize}
【详解】\\
$\bar{z}=-1-\sqrt{3} i, z \bar{z}=(-1+\sqrt{3} i)(-1-\sqrt{3} i)=1+3=4$.\\
$\frac{z}{z \bar{z}-1}=\frac{-1+\sqrt{3} \mathrm{i}}{3}=-\frac{1}{3}+\frac{\sqrt{3}}{3} \mathrm{i}$\\
故选: C\\
\\
The detailed solution for the given problem is as follows:\\
$\bar{z}=-1-\sqrt{3} i, z \bar{z}=(-1+\sqrt{3} i)(-1-\sqrt{3} i)=1+3=4$.\\
$\frac{z}{z \bar{z}-1}=\frac{-1+\sqrt{3} \mathrm{i}}{3}=-\frac{1}{3}+\frac{\sqrt{3}}{3} \mathrm{i}$\\
Therefore, the correct option is C.
\end{tcolorbox}


\subsection{Answer Example}
\label{app:ans_example}
An example of the answer of a geography open-ended question given by ChatGPT is shown below. 

\begin{tcolorbox}[breakable,colback=black!5!white,colframe=black!75!black]
\begin{itemize}
    \item \textbf{Prompt \& Question}
\end{itemize}
请解答下面的地理解答题。仔细阅读题目并充分结合你已有的知识，解答其中的问题，请你一步步思考并将思考过程写在【解析】和<eoe>之间。你的答案请写在【答案】和<eoa>之间。\\
完整的题目回答格式如下：\\
(1)【解析】 ...<eoe>【答案】...<eoa>\\
(2)【解析】 ...<eoe>【答案】...<eoa>\\
请你严格按照上述格式作答，如果不止一道题，请分别作答。\\
题目如下：\\
\\
位于陕西省安康市汉阴县的凤堰古梯田，是目前已发现的秦巴山区面积最大、保存最完整的清代梯田。 在总面积约 800 公顷的古梯田保护区内，分布有自清乾隆年间以来修建的古建筑民居群落、古寨堡、古庙宇、古堰渠、古塘坝等文物遗存。2019 年，凤堰古梯田被列为全国重点文物保护单位。以古梯田为“展品”， 以秀美的自然风光为“展厅”，凤堰古梯田被打造成保护和展示原生态生活方式、促进村镇科学发展的开 放式“生态博物馆”。提出该开放式“生态博物馆”开展旅游活动的旅游资源保护措施。\\
\\
Please answer the following geography essay question. Read the question carefully and fully integrate your existing knowledge to answer the questions. Think step by step and write your thought process in the section between 【Analysis】 and <eoe>. Write your answer in the section between 【Answer】 and <eoa>. \\
The complete format for answering the question is as follows: \\
(1)【Analysis】...<eoe>【Answer】...<eoa> \\
(2)【Analysis】...<eoe>【Answer】...<eoa> \\
Please strictly follow the above format to answer. If there is more than one question, please answer them separately. \\
The question is as follows:\\
\\
The Fengyan Ancient Terraces, located in Hanyin County, Ankang City, Shaanxi Province, are the largest and most well-preserved Qing Dynasty terraces discovered in the Qinba Mountain area to date. Within the approximately 800-hectare ancient terrace conservation area, there are distributed cultural relics such as ancient residential buildings, ancient fortresses, ancient temples, ancient canals, and ancient dams, which have been constructed since the Qing Dynasty's Qianlong era. In 2019, the Fengyan Ancient Terraces were listed as a national key cultural relics protection unit. With the ancient terraces as the ‘exhibit’ and the beautiful natural scenery as the ‘exhibition hall’, the Fengyan Ancient Terraces have been transformed into an open 'Ecological Museum' that protects and displays the original ecological lifestyle and promotes the scientific development of villages and towns. Propose tourism resource conservation measures for conducting tourism activities in this open ‘Ecological Museum’\\

\begin{itemize}
    \item \textbf{Output of ChatGPT}
\end{itemize}
【解析】\\
凤堰古梯田是一处重要的文化遗产，被列为全国重点文物保护单位。为了保护这一遗产，同时也为了开展旅游活动，需要采取一些措施来保护旅游资源。\\
【答案】\\
1. 加强管理：对凤堰古梯田保护区内的游客进行管理，设置必要的警示标志，禁止破坏梯田、采摘植物等行为。同时，加强对古建筑民居群落、古寨堡、古庙宇、古堰渠、古塘坝等文物遗存的保护，防止游客在参观过程中对这些文物遗存造成损害。\\
2. 推广科普：在凤堰古梯田保护区内设置科普展板，向游客介绍梯田的历史、文化和生态环境，提高游客的文化素养和环保意识，减少游客对梯田的破坏。\\
3. 限流措施：为了保护梯田的生态环境，可以设置游客限流措施，限制游客进入梯田的范围和数量，减少游客对梯田的踩踏和破坏。\\
4. 建设配套设施：为了方便游客参观，可以在凤堰古梯田保护区内建设一些配套设施，如停车场、休息区、卫生间等，提高游客的参观体验。同时，这些设施也需要合理规划，避免对梯田的破坏。\\
\\
【Analysis】\\
The Fengyan Ancient Terraces are an important cultural heritage and have been listed as a national key cultural relics protection unit. In order to protect this heritage while also developing tourism activities, measures need to be taken to protect the tourism resources.\\
【Answer】\\
1. Strengthen Management: Manage the tourists within the Fengyan Ancient Terraces conservation area, set up necessary warning signs, and prohibit actions such as damaging the terraces or picking plants. At the same time, enhance the protection of cultural relics such as ancient residential buildings, ancient fortresses, ancient temples, ancient canals, and ancient dams to prevent visitors from damaging these relics during their visit. \\
2. Promote Science Education: Install educational boards within the Fengyan Ancient Terraces conservation area to introduce tourists to the history, culture, and ecological environment of the terraces. This will enhance the cultural literacy and environmental awareness of the tourists and reduce the damage to the terraces. \\
3. Implement Visitor Limitations: To protect the ecological environment of the terraces, visitor limitations can be set to restrict the range and number of tourists entering the terraces, thus reducing trampling and damage. \\
4. Develop Supporting Facilities: To facilitate visitor access, some supporting facilities such as parking lots, rest areas, and restrooms can be constructed within the Fengyan Ancient Terraces conservation area, enhancing the visitor experience. At the same time, these facilities should be planned judiciously to avoid damage to the terraces.
\end{tcolorbox}


\subsection{Model Grading Example}
\label{model_grading_example}
We show the grading process of GPT-4-turbo corresponding to the answer given by ChatGPT in Appendix \ref{app:ans_example}.

\begin{tcolorbox}[breakable,colback=black!5!white,colframe=black!75!black]
\begin{itemize}
    \item \textbf{Prompt}
\end{itemize}
你是一名高中地理老师，正在批改高考地理试卷。\\
请根据下面的【题目】、【分析过程】、【标准答案】、【分值】、【评分点】、【学生分析与答案】，对【学生分析与答案】进行判分并给出理由。\\
请注意【学生分析与答案】可能为空。如果有多道小题，分别给出每道小题的得分，并在最后给出最终得分。\\
输出格式为：【判分理由】...【得分】...	...【总分】...分。\\
其中【总分】直接给出这道题的最终分数，如【总分】5分，注意不要超过这道题的【分值】。\\
请严格对照【标准答案】和【评分点】中的得分点。\\
\\
You are a high school geography teacher currently grading a geography paper from the college entrance examination. \\
Please score the 'Student Analysis and Answer' based on the following sections: 'Question', 'Analysis Process', 'Standard Answer', 'Points', 'Scoring Criterion', and 'Student Analysis and Answer'. \\
Note that the 'Student Analysis and Answer' section might be empty. If there are multiple sub-questions, provide scores for each sub-question separately and give the total score at the end.\\
The output format should be: 【Scoring Rationale】...【Points Awarded】...【Total Points】... points. \\
The 【Total Points】 should directly present the final score for the question, for example, 【Total Points】 5 points. Be careful not to exceed the total points available for the question. \\
Please strictly compare against the ‘Standard Answer’ and the scoring points in the ‘Scoring Criterion’.

\begin{itemize}
    \item \textbf{Marking Criterion}
\end{itemize}
【答案】完善景区规划、依法保护生态环境，对古建筑等进行保护性修缮，尽可能保证新建设施与景区景 观相融合。倡导绿色旅游，普及旅游文化环境保护教育，提高游客对旅游资源环境保护的意识。制定景区 相关环境保护条例与惩罚制度，评定该“生态博物馆”的环境容量，对人口数量的容纳程度，限制客流量。（任答三点10分）\\
\\
【Answer】Improve the planning of the scenic area, protect the ecological environment in accordance with the law, and carry out conservation repairs on ancient buildings to ensure that new facilities blend harmoniously with the scenic landscape. Advocate for green tourism, popularize education on the protection of the tourism cultural environment, and raise tourists' awareness of the protection of tourism resources and environment. Develop environmental protection regulations and punishment systems for the scenic area, assess the environmental capacity of this ‘Ecological Museum’, and regulate the carrying capacity in terms of population, limiting the flow of visitors. (Any three points for 10 points.)\\

\begin{itemize}
    \item \textbf{Output of GPT-4-turbo}
\end{itemize}
【判分理由】\\
1. 加强管理：学生提出了对游客进行管理和设置警示标志，以及加强对文物遗存的保护，这与标准答案中的“完善景区规划、依法保护生态环境”和“制定景区相关环境保护条例与惩罚制度”相符，可以给分。【得分】3分 \\
2. 推广科普：学生提出了设置科普展板来提高游客的文化素养和环保意识，这与标准答案中的“普及旅游文化环境保护教育，提高游客对旅游资源环境保护的意识”相符，可以给分。【得分】3分 \\
3. 限流措施：学生提出了设置游客限流措施，这与标准答案中的“评定该‘生态博物馆’的环境容量，对人口数量的容纳程度，限制客流量”相符，可以给分。【得分】2分 \\
4. 建设配套设施：学生提出了建设配套设施并合理规划，这与标准答案中的“尽可能保证新建设施与景区景观相融合”相符，可以给分。【得分】2分\\
【总分】10分\\
\\
【Scoring Rationale】\\
1. Strengthen Management: The student proposed managing tourists and setting up warning signs, as well as enhancing the protection of cultural relics, which aligns with the standard answer's "improve the planning of the scenic area, protect the ecological environment in accordance with the law" and "develop environmental protection regulations and punishment systems for the scenic area." Points can be awarded for this. 【Points Awarded】3 points \\
2. Promote Science Education: The student proposed setting up educational boards to enhance tourists' cultural literacy and environmental awareness, which aligns with the standard answer's "popularize education on the protection of the tourism cultural environment, raise tourists' awareness of the protection of tourism resources and environment." Points can be awarded for this. 【Points Awarded】3 points \\
3. Implement Visitor Limitations: The student proposed visitor limitation measures, which align with the standard answer's "assess the environmental capacity of this ‘Ecological Museum’, regulate the carrying capacity in terms of population, limit the flow of visitors." Points can be awarded for this. 【Points Awarded】2 points \\
4. Develop Supporting Facilities: The student proposed constructing supporting facilities and planning them reasonably, which aligns with the standard answer's "ensure new facilities blend harmoniously with the scenic landscape." Points can be awarded for this. 【Points Awarded】2 points \\
【Total Points】10 points

\end{tcolorbox}

\section{Converted Total Scores for Each Subject}
\label{app:cvt_score}
The converted total score for each subject is shown in Table \ref{tab:cvt_score}. We refer to the proportional weighting of scores for each subject in the GAOKAO over the years.

\renewcommand{\arraystretch}{1}
\begin{table*}[!htb]
\centering
\resizebox{\textwidth}{!}{
\begin{tabular}{lccc}
\hline
\textbf{Subject}   & \textbf{Objective Score} & \textbf{Subjective Score} & \textbf{Converted Total Score} \\
\hline
\textbf{Chinese}& 45& 105&150             \\ 
\textbf{English}& 105& 45&150             \\ 
\textbf{Sciences Mathematics}& 60& 90&150             \\ 
\textbf{Humanities Mathematics}& 60& 90& 150            \\ 
\textbf{Physics}& 44& 66& 110            \\ 
\textbf{Chemistry}& 50& 50& 100            \\ 
\textbf{Biology}& 27& 63& 90            \\ 
\textbf{Politics}& 50& 50& 100            \\ 
\textbf{History}& 50& 50& 100            \\ 
\textbf{Geography}& 40& 60& 100            \\ 
\hline
\end{tabular}
}
\caption{Converted Total Score for Each Subject.}
\label{tab:cvt_score}
\end{table*}
\renewcommand{\arraystretch}{1} 

\section{Statistics of Question Types}
\label{app:statistics_question_type}
The differences between various question types are significant, as demonstrated by the Table \ref{tab: statistics_q_t}:

\renewcommand{\arraystretch}{1.2}
\begin{table*}[htb]
\centering
\resizebox{\textwidth}{!}{
\begin{tabular}{lccc}
\hline
\textbf{Question Type} & \textbf{Type} & \textbf{Number} & \textbf{Avg. Length}  \\
\hline


\textbf{Biology MCQs} &single choice & 150& 162.1 \\ 
\textbf{Chemistry MCQs} &single choice & 124& 385.6 \\ 
\textbf{Chinese Lang and Usage MCQs} &multi-question choice & 56& 431.6 \\ 
\textbf{Chinese Modern Lit} &multi-question choice & 29& 1872.6 \\ 
\textbf{English Cloze Test} &five out of seven & 26& 299.5 \\ 
\textbf{English Fill in Blanks} &multi-question choice & 30& 448.3 \\ 
\textbf{English MCQs} &single choice & 105& 26.2 \\ 
\textbf{English Reading Comp} &multi-question choice & 124& 400.9 \\ 
\textbf{Geography MCQs} &multi-question choice & 34& 340.1 \\ 
\textbf{History MCQs} &single choice & 287& 158.8 \\ 
\textbf{Hum. Math MCQs} &single choice & 218& 157.4 \\ 
\textbf{Physics MCQs} &multi choice & 64& 285.5 \\ 
\textbf{Political Science MCQs} &single choice & 320& 228.9 \\ 
\textbf{Sci. Math MCQs} &single choice & 214& 181.6 \\

\hline


\textbf{Biology Open-ended Questions} &open-ended question & 116& 359.0 \\ 
\textbf{Chemistry Open-ended Questions} &open-ended question & 9& 1144.7 \\ 
\textbf{Chinese Language Ancient Poetry Reading} &open-ended question & 29& 310.2 \\ 
\textbf{Chinese Language Classical Chinese Reading} &open-ended question & 29& 1526.2 \\ 
\textbf{Chinese Language Famous Passages and Sentences Dictation} &cloze & 28& 219.5 \\ 
\textbf{Chinese Language Language and Writing Skills Open-ended Questions} &open-ended question & 42& 279.4 \\ 
\textbf{Chinese Language Literary Text Reading} &open-ended question & 29& 2222.6 \\ 
\textbf{Chinese Language Practical Text Reading} &open-ended question & 24& 2069.5 \\ 
\textbf{English Language Cloze Passage} &cloze & 23& 198.0 \\ 
\textbf{English Language Error Correction} &correction & 26& 120.2 \\ 
\textbf{Geography Open-ended Questions} &open-ended question & 28& 173.8 \\ 
\textbf{Hum. Math Fill-in-the-Blank} &cloze & 86& 97.7 \\ 
\textbf{Hum. Math Open-ended Questions} &open-ended question & 122& 273.6 \\ 
\textbf{Physics Open-ended Questions} &open-ended question & 47& 281.5 \\ 
\textbf{Political Science Open-ended Questions} &open-ended question & 60& 528.4 \\ 
\textbf{Sci. Math Fill-in-the-Blank} &cloze & 81& 112.2 \\ 
\textbf{Sci. Math Open-ended Questions} &open-ended question & 123& 273.8 \\

\hline
\end{tabular}
}
\caption{\label{tab: statistics_q_t}\textbf{Statistics of Question Types.} Questions above the line are objective questions, while questions below the line are subjective questions.}

\end{table*}
\renewcommand{\arraystretch}{1} 

\end{CJK*}

\end{document}